\documentclass[letterpaper, 10 pt, conference]{ieeeconf}  

\IEEEoverridecommandlockouts                              
\usepackage{graphicx} 
\usepackage{subfigure} 
\usepackage{url}
\usepackage{wrapfig}

\title{\LARGE \bf Human operator cognitive availability aware Mixed-Initiative control}

\author{Giannis Petousakis$^{1}$, Manolis Chiou$^{2}$, Grigoris Nikolaou$^{1}$, and Rustam Stolkin$^{2}$ 
\thanks{$^{1}$University of West Attica, Greece 
        {\tt\small gspetou@gmail.com, nikolaou@uniwa.gr}}%
\thanks{$^{2}$Extreme Robotics Lab (ERL) and National Center for Nuclear Robotics (NCNR), University of Birmingham, UK
        {\tt\small \{m.chiou, r.stolkin\}@bham.ac.uk}}%
}

\begin{document}

\maketitle
\thispagestyle{empty}
\pagestyle{empty}




\begin{abstract}

This paper presents a Cognitive Availability Aware Mixed-Initiative Controller for remotely operated mobile robots. The controller enables dynamic switching between different levels of autonomy (LOA), initiated by either the AI or the human operator. The controller leverages a state-of-the-art computer vision method and an off-the-shelf web camera to infer the cognitive availability of the operator and inform the AI-initiated LOA switching. This constitutes a qualitative advancement over previous Mixed-Initiative (MI) controllers. The controller is evaluated in a disaster response experiment, in which human operators have to conduct an exploration task with a remote robot. MI systems are shown to effectively assist the operators, as demonstrated by quantitative and qualitative results in performance and workload. Additionally, some insights into the experimental difficulties of evaluating complex MI controllers are presented.

\end{abstract}

\section{Introduction}
In this paper we follow a Variable Autonomy (VA) approach in order to improve Human-Robot Systems (HRS) that are being increasingly used in high risk, safety-critical applications such as disaster response. VA in this context refers to systems with autonomous capabilities of varying scales.

The task of manually controlling the robot combined with various operator straining factors, increase the cognitive workload of the operator \cite{CasperMurphy911}. Moving towards robotic systems that can actively assist the operators and reduce their workload can lead in reduced errors and improved performance \cite{CasperMurphy911,Yanco2015DARPA}. The advantage of such HRS lies in the complementing capabilities of humans and Artificial Intelligence (AI). In a VA system control can be traded between a human operator and a robot by switching between different Levels of Autonomy \cite{Chiou2016_IROS}, e.g. the robot can navigate autonomously while the operator is multitasking. Levels of Autonomy (LOAs) refer to the degree to which the robot, or any artificial agent, takes its own decisions and acts autonomously \cite{Sheridan1978}. This work addresses the problem of dynamically changing LOA during task execution using a form of VA called Mixed-Initiative control. Mixed-initiative (MI) refers to a VA system where both the human operator and the robot's AI have authority to initiate actions and to change the LOA \cite{Jiang2015}.

In many VA systems found in literature the LOA is chosen during the initialization of the system and cannot change on the fly \cite{Nielsen2008, Bruemmer2005}. Other systems allow only the operator to initiate actions, although these are based on system's suggestions  \cite{Gateau2016}. Hence, the agents lack the ability to override each other's actions. Additionally, the robot's AI initiative is often limited within a specific LOA, e.g. in safe mode \cite{Bruemmer2005}. Research on MI systems that are able to switch LOA dynamically is fairly limited. Moreover, some of the MI systems proposed are not experimentally evaluated e.g. \cite{Bruemmer2003b}.

Our previous work \cite{Chiou2019_arXiv} proposed a novel AI "expert-guided" MI controller and identified some major challenges: the conflict for control between the operator and the robot's AI and the design of context aware MI controllers. In this paper we extend the MI controller proposed in \cite{Chiou2019_arXiv} in order to tackle the above mentioned challenges by using information on the cognitive availability of the human operator and by improving the controller's assumptions. We call our approach Cognitive Availability Aware Mixed-Initiative (CAA-MI) control. Cognitive availability indicates if the operator's attention is available to focus on controlling the robot.

In our work, cognitive availability inference is based on the operator's head pose estimation provided by a state-of-the-art deep learning computer vision algorithm. A low cost, off-the-self webcamera mounted on the Operator Control Unit (OCU) provides video streaming to the computer vision algorithm in real time. The head pose estimation provides input to the fuzzy CAA-MI controller where the cognitive availability inference and the LOA switching decisions are made. Somewhat related to our paper is the work of Gateau et al. \cite{Gateau2016} which uses cognitive availability to inform decisions on asking the operator for help. Gateau et al. \cite{Gateau2016} uses a specialized eye tracker and does not involve cognitive availability informed LOA switching.

This work is explicitly making use of operator's cognitive availability in MI controller's LOA switching decision process and is contributing by: a) extending the MI controller in \cite{Chiou2019_arXiv} to explicitly include cognitive availability and active LOA information; b) including those parameters provides a qualitative advancement that tackles assumptions of the original controller; c) providing proof of concept on using state-of-the-art computer vision methods for informing MI control on operator's status.

\section{Expert-guided Mixed-Initiative control with cognitive availability inference}

The MI control problem this paper addresses is allowing dynamic LOA switching by either the operator or the robot's AI towards improving the HRS performance. In this work we assume a HRS which has two LOAs: a) teleoperation, where the human operator has manual control over the robot via a joystick; b) autonomy, where the operator clicks on a desired location on the map with the robot autonomously executing a trajectory to that location.

\begin{figure}
	\centering
	\includegraphics[width=0.7\columnwidth]{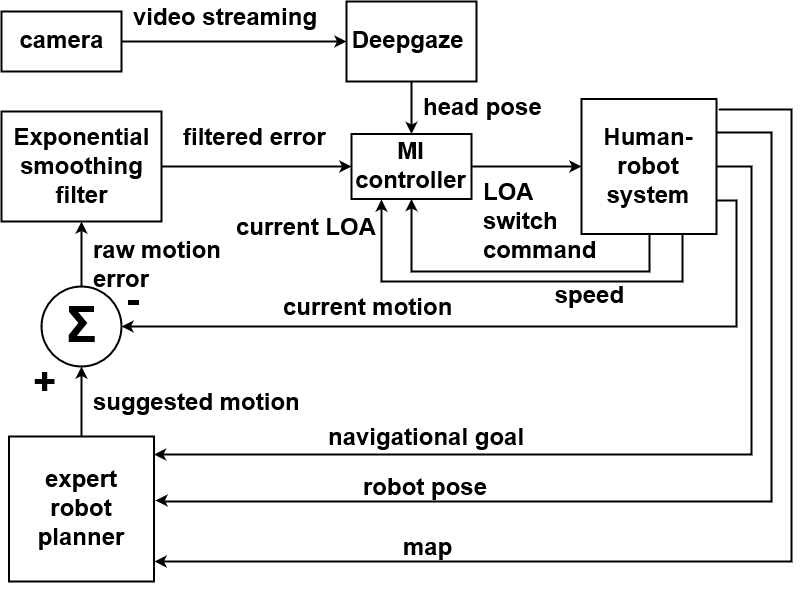}
	\caption{The block diagram of the CAA-MI control system. 
	} 
	\label{fig:block_diagram}
\end{figure}

Operator's initiated LOA switches are based on their judgment. Previous work \cite{Chiou2016_IROS,Chiou2016_AAAI} showed that humans are able to determine when they need to change LOA in order to improve performance. AI initiated LOA switches are based on a fuzzy inference engine. The controller's two states are: a) switch LOA; b) do not switch LOA. The controller uses four input variables: a) an online task effectiveness metric for navigation, the \textit{goal-directed motion error}; b) the cognitive availability information provided via the deep learning computer vision algorithm; c) the currently active LOA; d) the current speed of the robot. Please refer to Figure \ref{fig:block_diagram} for the block diagram of the system.

\subsection{Goal-directed motion error}

The CAA-MI controller uses an expert-guided approach to initiate LOA switches. It assumes the existence of an AI task expert that given a navigational goal can provide the expected task performance. The comparison between the run-time performance of the system with the expected expert performance yields an online task effectiveness metric. This metric expresses how effectively, the system, performs the navigation task. 

The controller's task effectiveness metric is the \textit{goal-directed motion error} (refereed to as ``error'' from now on) and is the difference between the robot's current motion (i.e. speed in this case) and the motion of the robot required to achieve its goal according to an expert. To provide more context we extract the expert motion performance from a concurrently active navigation system which is given an idealized (unmapped obstacle and noise-free) view of the robot's world. In essence, the expert provides an idealized model of possible robot behavior that can be seen as an upper bound on system performance.

Additionally, our controller encodes expert knowledge from human operators data learned using machine learning techniques from a previous experiment \cite{Chiou2016_IROS} on: a) what is considered to be a large enough error to justify a LOA switch; and b) the time window in which that error needs to accumulate. For detailed information the reader is encouraged to read our previous work \cite{Chiou2019_arXiv}.

\subsection{Cognitive availability via head pose estimation}

In this work we consider the information on whether the operator is attending or not at the Graphical User Interface (GUI) to be directly relevant to his current cognitive availability. The CAA-MI controller presented here infers the operator’s cognitive availability by monitoring the operator’s head pose via a computer vision algorithm. By introducing cognitive availability perception capabilities to the HRS system, we make a qualitative change to the robot's ability to perceive information about the operator and his availability to control the robot.

By using a computer vision technique for inferring the cognitive availability of the operator we avoid intrusive methods such as biometric sensors (e.g. EEG) and the use of cumbersome apparatus. Our aim is to keep the approach generalizable and versatile. The latest head pose estimation techniques are robust in real time use and do not require specialized image capturing equipment \cite{Patacchiola2017,Ruiz2018}, contrary to most state-of-the-art gaze tracking implementations (e.g. \cite{Lemley2019}). We chose head pose estimation over gaze tracking preferring perception of more pronounced natural cues. 

Our focus is on using a robust CV algorithm to provide cognitive availability input to the controller rather than evaluating the various CV algorithms. The Deepgaze algorithm \cite{Patacchiola2017} is used in our system for head pose estimation. It makes use of state-of-the-art methods such as a pre-trained Convolutional Neural Network (CNN), is open source and actively maintained. In our implementation, deepgaze measures the operator's head rotation on the vertical axis (yaw). The head rotation was filtered through an exponential moving average (EMA) \cite{Brown1963} in order to reduce noise in the measurements. The parameters of the EMA were calculated by analyzing data regarding the attention of the operator in the secondary task in previous experiments. Lastly, baseline data on what values of rotation constitute attending or not at the GUI were gathered in a pilot experiment and in a standardized way. These were mapped into the fuzzy input for cognitive availability.

\subsection{The fuzzy rule base}
A fuzzy bang-bang controller is used with the Largest of Maxima defuzzification method. We have introduced the active LOA and the Cognitive availability of the operator as fuzzy inputs. A hierarchical fuzzy approach is used for the CAA-MI controller, meaning that the first in order rule activated has priory over the others. The cognitive availability of the operator and the active LOA take precedence in the fuzzy rule activated. The rules added to the original MI controller make sure that when the operator is not attending the screen the robot will operate in autonomy, with the rest of the rule base architecture being similar to the previous MI controller.  

\section{Experimental evaluation}
\begin{figure}
	\centerline{\subfigure[]{\includegraphics[width=0.41\columnwidth]{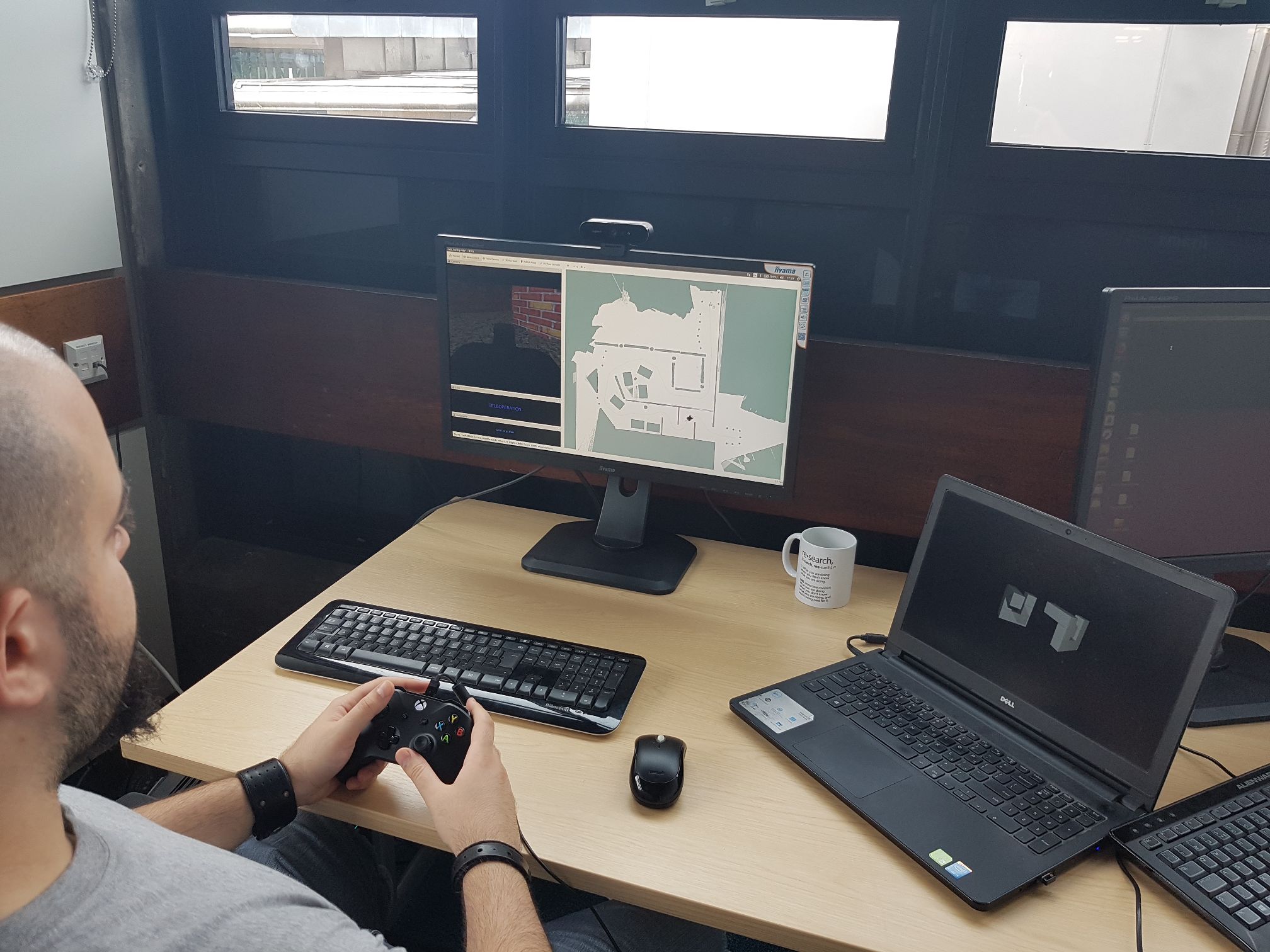}
			\label{fig:ocu}}
		\hfil
		\subfigure[]{\includegraphics[width=0.42\columnwidth]{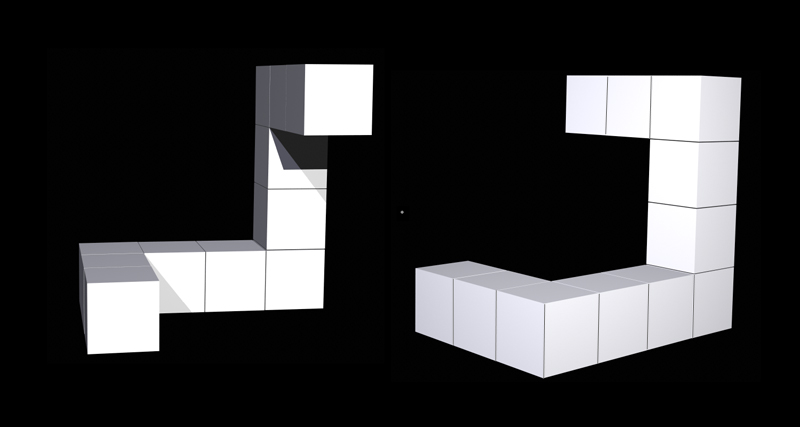}
			\label{fig:secondary_example}}}
	\caption{\textbf{\ref{fig:ocu}:} The experimental apparatus: composed of a mouse and a joystick, a laptop and a desktop computer, a screen showing the GUI; a web-camera mounted on the screen; and a laptop presenting the secondary task. \textbf{\ref{fig:secondary_example}:} A typical example of the secondary task.}
	\label{fig:ocu_and_secondary_image}
\end{figure}

\begin{figure}
	\centering
	\includegraphics[width=0.95\columnwidth]{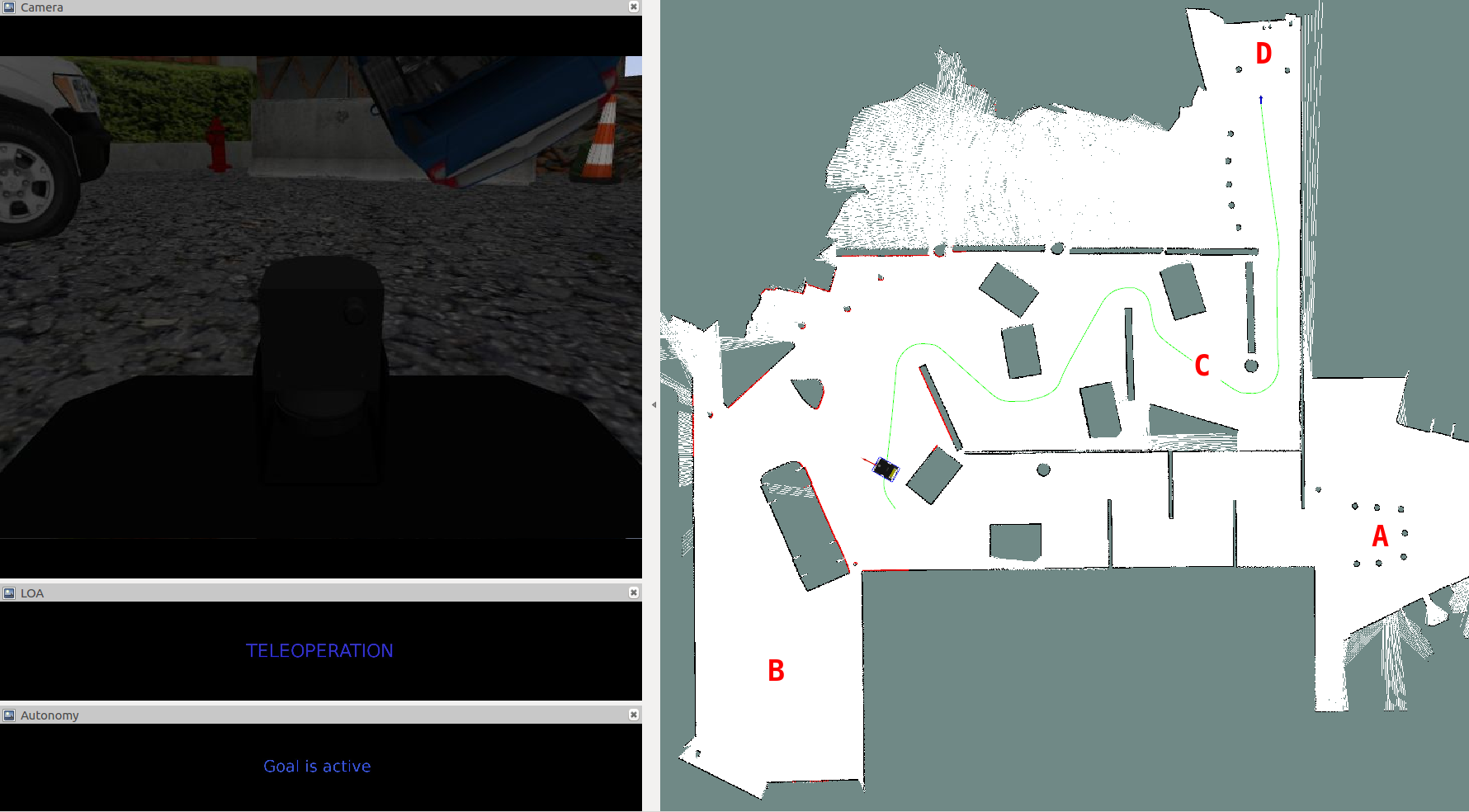}
	\caption{The GUI: \textbf{left}, video feed from the camera, the LOA in use and the status of the robot; \textbf{right}, the map showing the position of the robot, the current goal (blue arrow), the AI planned path (green line), the obstacles’ laser reflections (red) and the walls (black). The letters denote the waypoints to be explored.} 
	\label{fig:gui}
\end{figure}

This experiment evaluates the CAA-MI controller and investigates its potential advantages over the previously built MI controller \cite{Chiou2019_arXiv} (referred simply as MI controller for the rest of the paper). The Gazebo simulated robot was equipped with a laser range finder and a RGB camera. The software was developed in Robot Operating System (ROS) and is described in more detail in \cite{Chiou2019_arXiv}. The ROS code for the CAA-MI controller is provided under MIT license in our repository \cite{Chiou_Petousakis_code}. The robot was controlled via an Operator Control Unit (OCU) (see Figure \ref{fig:ocu}) including a GUI (Figure \ref{fig:gui}). The experiment's test arena was approximately $24m \times 24m$. The software used for the secondary task was the OpenSesame \cite{Mathot2012OpenSesame} and the images used as stimuli were previously validated for mental rotation tasks in \cite{Ganis2015}.

The participants were tasked with a primary exploration task and a cognitively demanding secondary task. The primary task was the exploration of the area along a set of waypoints in order to identify the number of cardboard boxes placed on those waypoints. In the secondary task the operators were presented with pairs of 3D objects. The operators were required to verbally state whether or not the two objects were identical or mirrored.
 
Artificially generated sensor noise was used to degrade the performance of the autonomous navigation. The secondary task was used to degrade the performance of the operator. Each of these performance degrading situations occurred once concurrently (i.e. overlapping) but at random. 

A total of 8 participants participated in a within-groups design in which every participant performed two identical trials, one with each controller: the CAA-MI controller and the MI controller. Counterbalancing was used in the order of the two controllers for different participants. Each participant underwent extensive standardized training ensuring a minimum skill level and understanding of the system. 
Participants were instructed to perform the primary task as quickly and safely as possible. They were also instructed that when presented with the secondary task, they should do it as quickly and as accurately as possible. They were explicitly told that they should give priority to the secondary task over the primary task and should only perform the primary task if the workload allowed. Lastly, after every trial, participants completed a raw NASA-TLX workload questionnaire.

\section{Results}

\subsection{Statistical analysis}

 The CAA-MI controller displayed a trend, though not statistically significant (Wilcoxon signed-rank tests and t-tests), of improved performance over the MI controller in terms of lower number of LOA switches, more secondary task number of correct answers and less perceived workload (NASA-TLX), as demonstrated in table \ref{table:results}.

\begin{table}[ht]
\caption{Table showing descriptive statistics for all the metrics.}
	\centering
	\begin{tabular}{ll}

		\hline
		\textbf{metric}                                 & \textbf{descriptive statistics}                                                                                                                               \\ \hline
		\begin{tabular}[c]{@{}l@{}}primary task\\ completion time\end{tabular}    &  \begin{tabular}[c]{@{}l@{}}MI: $M = 242$ $sec$, $SD = 21.8$ \\CAA-MI: $M = 243.5$ $sec$, $SD = 24.4$ \end{tabular} \\ \hline
		
		\begin{tabular}[c]{@{}l@{}}secondary task no.\\of correct answers\end{tabular} &  \begin{tabular}[c]{@{}l@{}}MI: $M = 7.6$, $SD = 3.1$ \\ CAA-MI: $M = 8.5$, $SD = 3.5$\\ Baseline: $M = 8.6$, $SD = 3.1$\end{tabular}       \\ \hline

		\begin{tabular}[c]{@{}l@{}}secondary task \\ accuracy (\% of\\correct answers) \end{tabular}          &  \begin{tabular}[c]{@{}l@{}}MI: $M = 84.2$, $SD = 15.6$ \\ CAA-MI: $M = 87.2$, $SD = 18.9$ \\ Baseline: $M = 82.9$, $SD = 14.2$\end{tabular}                          \\ \hline
		\begin{tabular}[c]{@{}l@{}}NASA-TLX\\(Lower score means\\ less workload)\end{tabular}                 &  \begin{tabular}[c]{@{}l@{}}MI: $M = 34.9$, $SD = 17.4$ \\ CAA-MI: $M = 30.4$, $SD = 14.97$ \end{tabular}                       \\ \hline
		\begin{tabular}[c]{@{}l@{}}number of \\ LOA switches\end{tabular}                                 & \begin{tabular}[c]{@{}l@{}}MI: $M = 6.4$, $SD = 3.8$ \\ CAA-MI: $M = 5.5$, $SD = 3.4$\end{tabular}           \\ \hline
		\begin{tabular}[c]{@{}l@{}}number of AI \\ LOA switches\end{tabular}         & \begin{tabular}[c]{@{}l@{}}MI: $M = 2$, $SD = 1.6$ \\ CAA-MI: $M = 1.75$, $SD = 1.5$\end{tabular}           \\ \hline
	\end{tabular}
	\label{table:results}
\end{table}

\subsection{Discussion}

Performance towards the primary task was on the same level for both controllers. This can be attributed to the varied exploration strategies of the operators as supported by anecdotal evidence. Additionally, it is possible that the performance degradation periods did not last enough to amplify the advantages of the CAA-MI controller compared to the MI one. By increasing secondary task duration and complexity, clearer insight to the advantages of the CAA-MI controller can be provided. Given the above evidence, a trade-off must be found between having a meaningful exploration task (i.e. to be used for system evaluation) and minimizing variance in operator strategies. 

During MI controller trials and while the secondary task overlapped with the sensor noise, the robot would switch from autonomy to teleoperation. This negatively affected operators' focus on the secondary task with some of them expressing their frustration. In contrast, this was not observed in the case of the CAA-MI controller since the robot could perceive that the operator was unavailable and hence maintained, or switched to, autonomy. Supporting evidence can be found in the NASA-TLX frustration scale which while not statistically significant, showed lower frustration in CAA-MI ($M = 27.5, SD = 21.9$) compared to MI ($M = 33.25, SD = 18.3$). Participants also performed better on the secondary task. In conjunction with the above, a trend for lower LOA switches has been observed in the case of the CAA-MI controller. This is a possible indication of more efficient LOA switching compared to the MI as the operators relied more on the CAA-MI system's capabilities.

Lastly, note that our previous work \cite{Chiou2016_IROS,Chiou2019_arXiv} has already demonstrated that dynamic LOA switching (e.g. MI control) is significantly better than pure teleoperation or/and pure autonomy. Hence, it can be inferred that the CAA-MI controller is also advantageous over pure teleoperation or pure autonomy.

\section{Conclusions and future work}

This paper presented a Cognitive Availability Aware Mixed-Initiative (CAA-MI) controller and its experimental evaluation. The CAA-MI controller's advantage lies with the use of operator's cognitive availability status into the AI's LOA switching decision process.

Research in MI control often requires integrating different advanced algorithms in robotics and AI. Thus, the use of a state-of-the-art deep learning computer vision algorithm, demonstrated proof of concept that such algorithms can provide reliable and low cost advancements to MI control.

Experimental results showed that the CAA-MI controller performed at least as good as the MI controller in the primary exploration task. Also trends found in the results (e.g. NASA-TLX frustration scale, number of LOA switches) and anecdotal evidence indicate a more efficient LOA switching in certain situations compared to the MI controller. This is due to the qualitative advantage that the operator's cognitive status information gives to the CAA-MI compared to the MI controller.

Concluding, we identify the current experimental paradigms as a limitation in testing the full potential of more complex MI systems. This is despite the recent advances on experimental frameworks \cite{Chiou2016_IROS}. Evaluating MI Human-Robot Systems in realistic scenarios involves difficult intrinsic confounding factors that are hard to predict or overcome. Hence, further research should devise a more complex experimental protocol along the lines proposed in the discussion, i.e. finding a trade-off between realism and meaningful scientific inference.

\section*{Acknowledgment}
This work was supported by the following grants of UKRI-EPSRC: EP/P017487/1  (Remote  Sensing  in  Extreme Environments); EP/R02572X/1 (National  Centre  for  Nuclear  Robotics); EP/P01366X/1 (Robotics for  Nuclear Environments). Stolkin was also sponsored by a Royal Society Industry Fellowship.

\bibliography{IEEEabrv,refs}
\bibliographystyle{IEEEtran}

\end{document}